\documentclass[double]{article}
\usepackage{times}
\usepackage{algorithm}
\usepackage{wrapfig}
\usepackage{verbatim}
\usepackage{amsmath}
\usepackage{graphicx}
\usepackage{subcaption}
\usepackage[active]{srcltx}
\usepackage{color}
\usepackage{lineno}
\usepackage{dirtytalk}
\usepackage{amsthm}
\usepackage{authblk}
\usepackage{listings}
\usepackage{tikz}
\usepackage{longtable}
\usetikzlibrary{shapes.geometric, arrows}

\usepackage{float}
\usepackage{placeins}
\usepackage{booktabs}

\usepackage{epsfig,graphicx,amsfonts}
\usepackage{amsmath,amsthm,amssymb}
\usepackage{xcolor}

\usepackage{adjustbox}
\usepackage{multirow}
\usepackage{setspace}
\usepackage{hyperref}
\usepackage{comment}

\usepackage{algpseudocode}

\long\def\comment #1\commentend{}

\begin{document}

\title{\Large Data Enrichment for Symbolic Regression Using Diffusion Models}

\author{Simon De Reuver$^{1}$, Tamás Kristóf Tóth$^{1}$, Teddy Lazebnik$^{1,2,*}$ \\ \(^1\)  Department of Computing, Jönköping University, Jönköping, Sweden \\ \(^2\) Department of Information Science,  University of Haifa, Haifa, Israel\\ \(^*\) Corresponding author: teddy.lazebnik@ju.se }
%\author{Anonymous for Review}

\date{ }

\maketitle 

\begin{abstract}
\noindent
Symbolic regression (SR) offers a route to scientific discovery by converting observations into interpretable governing equations. However, despite its promise, its reliability degrades sharply when spatiotemporal measurements are sparse, noisy, or physically incomplete, as commonly occurring in practice. Data enrichment (DE) has been shown to be able to mitigate this limitation, yet additional samples can mislead equation discovery unless they preserve the physical structure of the target system. Such implication of DE requires narrow domain expertise as well as technical fluidity, highly limiting its practical usefulness. In this study, we introduce a physics-guided latent diffusion framework for DE for down the line SR models. The proposed framework combines a variational autoencoder, a conditional latent diffusion model, and a physics-informed residual corrector to complete sparse observations with synthetic fields constrained by governing relations. We evaluate the approach on heat conduction, incompressible Navier–Stokes flow, and a moving single-mass Newtonian gravitational potential, using GPLearn, DEAP, and PySR as downstream SR backends. Our results reveal that physics-corrected enrichment consistently improves recovery in sparse regimes across physical dynamics and SR models. These results show that generative enrichment can strengthen equation discovery without additional domain expertise.

\end{abstract}

\maketitle \thispagestyle{empty}
% Begin using page numbers and a header
\pagestyle{myheadings} \markboth{Draft:  \today}{Draft:  \today}
\setcounter{page}{1}% reset page number to 1

\section{Introduction}
\label{sec:introduction}
Symbolic regression (SR) is a family of methods that predict closed-form mathematical expressions given empirical data \cite{wang2019symbolic,muthyala2025symantic}. Recently, SR has become a prominent route for data-driven scientific discovery due to its ability to produce models that are both predictive and interpretable \cite{shmuel2026interpretable,makke2024review,kim2020integration}. From the automated rediscovery of conservation laws and mechanical invariants to a broad and growing range of modern scientific applications, SR is valued precisely because it can turn data into equations that scientists can inspect, test, and reuse \cite{schmidt2009distilling,makke2024review}.

Despite this promise, SR remains acutely sensitive to the quality, density, and coverage of the available measurements, limiting its practical usefulness as empirical data suffers from data quality, density, and coverage \cite{vaddireddy2020feature,cohen2024pdegp}. This sensitivity is particularly severe in physical dynamical systems where observations are sparse, noisy, or irregularly sampled in time, because derivative estimates and trajectory constraints can become unreliable \cite{goyal2022rk4sindy,dascoli2024odeformer}. This phenomenon is well documented in multiple recent attempts of using SR models for such systems \cite{lu2022partial,egan2024argos}.

To this end, data enrichment (DE) can substantially improve SR models' performance when the original measurements are sparse or noisy \cite{hsin2024gpsindy,meng2024mfgp,zhao2023coarsenoisy}. For instance, Hsin et al.\ enriched limited time-series data by first fitting Gaussian processes and then using the smoothed trajectories and derivative estimates for downstream symbolic identification, showing that this preprocessing step made equation discovery markedly more robust under sparse and noisy sampling conditions \cite{hsin2024gpsindy}. Alike, Meng and Qiu extended this idea to the multi-fidelity setting, using Gaussian-process fusion to combine sparse, noisy, low- and high-fidelity observations into a more informative dataset for sparse equation discovery, and showed that this enriched support set leads to more accurate and robust recovery of differential equations \cite{meng2024mfgp}. In a similar manner, Zhao and Wang addressed coarse and noisy observations by learning a physics-informed surrogate of the dynamics before SR, effectively replacing poor-quality measurements with a denser and more structured representation of the system, which improved recovery on challenging chaotic and stiff dynamics \cite{zhao2023coarsenoisy}. These examples obtained downstream SR improvement from the data enrichment, as the synthetic data preserved the physical structure of the underlying law informed by the authors. The challenge, however, is that not all additional samples are necessarily helpful. For SR, synthetic data are valuable only if they expand coverage without corrupting the physical structure of the underlying law it aims to reconstruct \cite{huang2025physics,han2025physics}. In practice, generating such physically admissible samples is difficult, especially for spatiotemporal systems with limited observations and without deep domain expertise. 

In this context, recent advances in diffusion models (DM) suggest a promising path forward \cite{cui2025diffusionst,chen2024overview}. DMs are generative models that learn to produce new data samples by reversing a gradual noising process \cite{cao2024survey}. Thus, a DM can, in principle, enrich the dataset by generating additional samples drawn from the learned distribution of physically relevant trajectories \cite{gao2024bayesian,cechnicka2023realistic}. When supplemented with generally true physics-informed guidance, these generated samples can better preserve the structure of the underlying system, thereby improving the support available for downstream equation discovery. Indeed, physics-aware DMs have been used to reconstruct high-fidelity flow fields from sparse or low-fidelity inputs, to directly enforce PDE consistency during diffusion training, and to generate rich spatiotemporal turbulence in latent space under varied conditions \cite{shu2023flowdiff,bastek2025pidm,du2024confield}.

In this study, we build on this line of work and investigate whether and to what extent physics-guided diffusion can serve as a data-enrichment mechanism for SR. We propose a framework that combines a variational autoencoder (VAE), a latent diffusion model (LDM), and a physics-informed corrector in order to complete sparse spatiotemporal observations with synthetic samples that are both distributionally plausible and physically constrained. We evaluate the proposed framework across several popular out-of-the-box SR models and on three classical spatio-temporal physical problems. 

The rest of the paper is organized as follows. Section~\ref{sec:rw} reviews related work on SR, data-centric equation discovery, and physics-aware generative modeling. Section~\ref{sec:model} presents the proposed framework. Section~\ref{sec:experiments} describes the experimental design, including the physical systems, data-generation procedures, SR baselines, sparsity protocol, and evaluation metrics. Section~\ref{sec:results} reports the obtained empirical evaluation. Finally, Section~\ref{sec:discussion} discusses the implications, limitations, and future directions of using generative enrichment as a complement to SR.

\section{Background}
\label{sec:rw}
In this section, we outline the three methodological pillars on which this study is based: SR, generative DMs, and physics-informed machine learning. 

\subsection{Symbolic regression} 
SR seeks to discover explicit mathematical expressions directly from data without fixing the functional form in advance \cite{dong2025recent}. In scientific settings, this makes SR attractive both as a predictive tool and as a mechanism for producing interpretable hypotheses about the underlying system \cite{angelis2023symbolic}. It is possible to reverse-engineer nonlinear dynamical systems directly from time-series data by searching over symbolic expressions, while later sparse-regression formulations recast equation discovery as the identification of a parsimonious subset of active terms from a large candidate library \cite{schaeffer2017learning,zhang2018robust}. This shift was especially influential because it connected SR-based discovery to dynamical-systems identification and to the recovery of governing ordinary differential equations (ODEs) and PDEs from measurements 
\cite{bongard2007automated,brunton2016discovering,rudy2017datadriven}.

SR methods can be broadly grouped according to the search strategy used to explore the space of symbolic expressions, including exhaustive or enumeration-based search, sparse-regression approaches, neural or deep-learning-guided methods, and evolutionary or genetic-programming-based methods \cite{kronberger2024symbolicregression}. The brute-force-based SR methods can theoretically solve many SR tasks under a properly restricted search space. Nevertheless, practically, their computational cost is unrealistic, and they are extremely prone to overfitting \cite{sr_23}. The method tests all possible equations to find the one that performs optimally, differing in the way they cover the search space \cite{sr_22}. The sparse regression SR method, which identifies parsimonious models with the help of optimization that promotes sparsity, significantly reduces the search space. Among them, SINDy is designed for scientific use cases, which employs the Lasso linear model for sparse identification of nonlinear dynamical systems behind time series data \cite{brunton2016discovering}. SINDy iterates between sequentially thresholded least squares steps \cite{sr_21}. Deep learning SR methods excel in handling noisy data, but are limited in their generality \cite{LaCava2021SRBench}. These methods use neural networks to produce analytical equations. For instance, Petersen et al. proposed a Deep SR (DSR) system for general SR tasks, utilizing reinforcement learning to train a generative recurrent neural network (RNN) model of symbolic expressions and employs a variant of the Monte Carlo policy gradient technique called \say{risk-seeking policy gradient} to adapt the generative model to the exact formulation \cite{sr_17}. Lastly, genetic-algorithm-based SR methods treat mathematical expressions as individuals (or genes) in a population \cite{sr_16}. These individuals evolve over time through mechanisms such as selection, crossover, and mutation, gradually refining the equations that can better fit the data. This evolutionary approach allows SR to discover interpretable models \cite{sr_4,sr_5}. For example, the gplearn Python library implements a genetic algorithm for SR, showing a promising ability to re-discover known physical equations \cite{sr_19}. It first constructs a population of stochastic formulas representing the relationship between known independent variables (features) and dependent variables (objectives), presented in a tree structure. These subtrees are then replaced and restructured in a stochastic optimization process that computes fitness by executing the tree and evaluating its output, using a stochastic strategy of survival of the fittest. In recent years, many SR algorithms have been proposed~\cite{tohme2022gsr,landajuela2022udsr,jiang2024vsrdpg}. From these, several SR tools have gained popularity and provide a simple programming interface to use, including DEAP \cite{fortin2012deap}, gplearn \cite{sr_19}, and PySR \cite{cranmer2023pysr}.

Notably, SR performance depends critically on the representation of the data, not just on the optimizer used to search over formulas \cite{cranmer2020discovering,matsubara2024rethinking}. For instance, Champion et al.\ addressed this issue by coupling representation learning with equation discovery, showing that one can first identify a lower-dimensional coordinate system and then discover governing equations in that learned space \cite{champion2019coordinates}. Kaheman et al.\ proposed SINDy-PI to handle implicit dynamics and rational nonlinearities more robustly than standard explicit sparse regression \cite{kaheman2020sindypi}. In a similar manner, Messenger and Bortz introduced Weak-SINDy, replacing pointwise differentiation by a weak formulation that transfers derivatives onto test functions, thereby improving robustness to noise \cite{messenger2021weaksindy}. In addition, Fasel et al.\ showed that ensemble strategies can stabilize sparse model selection in low-data regimes by aggregating discoveries across resampled datasets rather than trusting a single brittle solution path \cite{fasel2022ensemblesindy}.

In this context, the data dependence is especially severe for spatiotemporal systems \cite{stephany2024weakpdelearn}. When the target law contains multiple spatial derivatives, transport terms, or nonlinear interactions, finite-difference estimates become fragile under coarse sampling and noise \cite{stephany2024pdelearn}. To this end, Both et al.\ proposed DeepMoD, which combines a neural network representation of the underlying field with sparse regression over candidate PDE terms, allowing smoother derivative evaluation and improved recovery from noisy spatiotemporal data \cite{both2021deepmod}. Tang et al.\ later introduced WeakIdent, a weak-formulation method with narrow-fit and trimming mechanisms designed to improve robustness for both ODEs and PDEs under noisy conditions \cite{tang2023weakident}.

\subsection{Generative diffusion models}
DMs are generative models that learn a data distribution by reversing a gradual corruption process \cite{sohl2015deep,song2019generative,luo2022understanding}. Simply put, the training of such models starts with clean samples that are progressively perturbed until they approach pure noise, and a neural network is trained to approximate the reverse dynamics  \cite{voleti2022mcvd}. Score-based formulations placed this idea in a continuous-time stochastic differential equation framework and showed that reverse-time sampling, predictor-corrector samplers, and exact-likelihood variants can all be understood within one unified perspective \cite{song2021scorebased} \cite{anderson1982reverse,hyvarinen2005scorematching}. Correspondingly, improved denoising diffusion probabilistic models demonstrated that learned reverse variances can substantially reduce the cost of sampling while preserving sample quality, helping make diffusion practical beyond purely proof-of-concept settings \cite{nichol2021improved}.

A particularly useful aspect of such DMs is their ability to support conditioning and guidance \cite{dhariwal2021diffusion,ho2022classifierfree}. Rather than only generating samples unconditionally from noise, they can be driven toward outputs consistent with observed data or external objectives, which makes them useful to reconstruction and inverse problems, which are common across many fields \cite{song2022solving,kawar2022denoising}. For instance, Saharia et al.\ illustrated this ability for the super-resolution task (i.e., a task that gets an image and returns the same image but in higher resolution), where repeated refinement conditioned on low-resolution inputs produced high-fidelity reconstructions through iterative denoising \cite{saharia2021sr3}. Akin, Chung et al.\ extended the same intuition to general noisy inverse problems through diffusion posterior sampling, showing that pretrained diffusion priors can be coupled with the measurement process during sampling to solve a broad class of reconstruction problems without retraining a separate generator for each task \cite{chung2023diffusionposterior}. 

That capability is particularly relevant for scientific data, where the target object is often a field or trajectory rather than an isolated sample \cite{kovachki2023neuraloperator}. In many scientific inverse problems, sparse observations do not uniquely determine the underlying state, so a useful model should represent a distribution of plausible completions rather than a single deterministic interpolation  \cite{stuart2010inverse,asch2016dataassimilation}. To this end, Rozet and Louppe framed data assimilation in exactly this way by learning a score-based generative model over trajectories and then guiding inference with incomplete observations at test time \cite{rozet2023scorebasedda}. From the viewpoint of PDE-based spatiotemporal modeling, Shysheya et al.\ studied conditional and post hoc conditioned DMs trained on short PDE trajectory segments and evaluated them on both forecasting and data-assimilation tasks, arguing that diffusion offers a flexible route to handling sparse observations without retraining task-specific solvers \cite{shysheya2024conditional}. Similarly, Huang et al.\ proposed DiffusionPDE, a generative framework that jointly models solutions and coefficient fields so that forward and inverse PDE problems can be addressed under partial observation \cite{huang2024diffusionpde}. Taking into larger scale, Price et al.\ introduced GenCast, a machine-learning weather model that produces probabilistic forecasts and showed strong performance against leading operational baselines \cite{price2025gencast}. Notably, this study highlights the point that DMs can capture multimodal future evolution and structured spatial correlations in regimes where deterministic surrogates may be too rigid.

\subsection{Physics-informed machine learning}
Physics-informed machine learning (PIML) incorporates scientific structure into data-driven models by embedding governing equations, conservation laws, boundary conditions, constitutive relations, or other mechanistic constraints into the learning process \cite{vonrueden2023informed,karpatne2017theoryguided,debezenac2018physical}. Recent studies describe this as a broad methodological shift from purely data-fitting models toward models that learn under both observational and physical supervision \cite{karniadakis2021piml,cuomo2022scientific,willard2022integrating}. In parallel, case-study syntheses in weather and climate have shown that such hybridization is especially valuable when full-resolution simulations or dense observations are expensive, making physical knowledge a crucial complement to limited data \cite{kashinath2021case}.

Commonly, PIML is implemented by augmenting a data loss with one or more residual terms derived from the governing equations \cite{raissi2019physics,lagaris1998artificial,sirignano2018dgm}. Technically, if a field $u$ is expected to satisfy a differential operator $\mathcal{F}(u)=0$, then one penalizes violations of that operator at collocation points in space-time. Conceptually, these residual terms act as soft physical constraints by discouraging solutions that fit the measurements but contradict the known physics \cite{raissi2019physics,cuomo2022scientific}. The appeal of this strategy is most obvious in ill-posed inverse problems, where observational data alone do not uniquely identify the latent state. In such cases, physical structure acts as an additional source of information rather than merely as a preference for smoother outputs \cite{yang2021bpinns}.

Indeed, multiple studies show the usefulness of this idea \cite{tartakovsky2020physics,lu2021deepxde}. For example, Jagtap et al.\ introduced conservative Physics-Informed Neural Networks (PINNs) that enforce flux continuity across subdomains, thereby improving the treatment of conservation laws and sharpening the link between the learning architecture and the underlying PDE structure \cite{jagtap2020conservative}. Jin et al.\ developed NSFnets for the incompressible Navier--Stokes equations, illustrating how alternative mathematical formulations of the same governing system can be embedded into the training objective to improve flow modeling \cite{jin2021nsfnets}. Raissi et al.\ used physics-informed learning in an inverse setting through hidden fluid mechanics, showing that velocity and pressure fields can be inferred from flow visualizations by embedding Navier--Stokes constraints directly into the model \cite{raissi2020hidden}. 

Importantly, PIML's ability to stabilize inference under limited observations is directly relevant to reconstruction and enrichment tasks \cite{xu2023practical,hosseini2024flowfield}. Accordingly, Golden et al.\ provide an instructive example by building a physically informed data-driven model of active nematics from experiments, using physical structure to recover a complete mathematical description of a complex nonequilibrium material system \cite{golden2023activenematics}. Eusebi et al.\ similarly showed that sparse observations of tropical cyclones can be lifted into realistic two- and three-dimensional wind and pressure reconstructions using a physics-informed neural network \cite{eusebi2024tcyclone}. 

\section{Physics-Informed Diffusion-Based Data Enrichment Model For Symbolic Regression}
\label{sec:model}
\begin{figure}[!ht]
    \centering
    \includegraphics[width=0.99\linewidth]{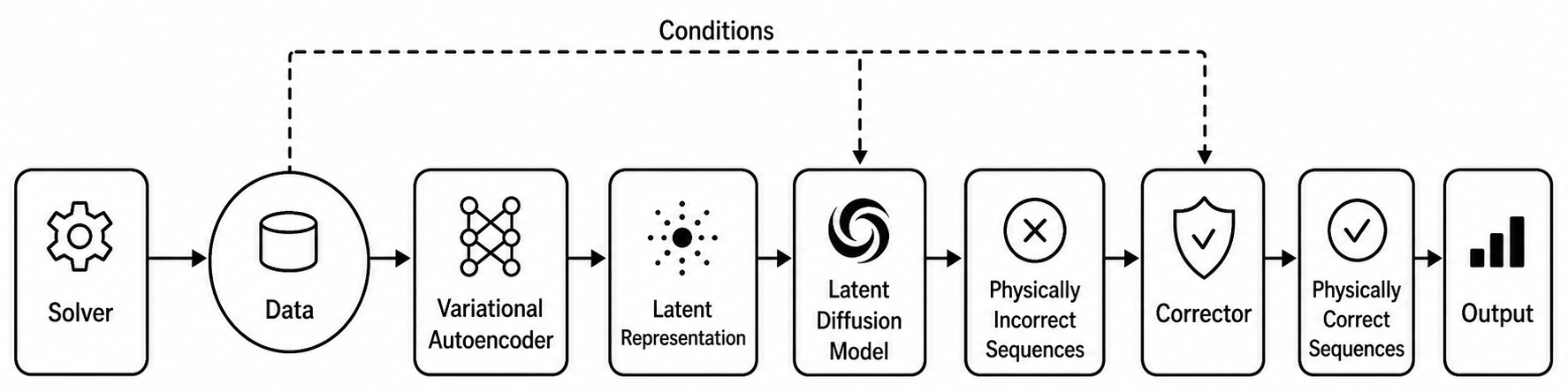}
    \caption{A schematic view of the proposed method.}
    \label{fig:scheme}
\end{figure}

In this study, we formally introduce the proposed framework, which consists of three components: a VAE, a LDM, and a physics corrector. The VAE maps observed spatiotemporal fields into a compact latent representation and reconstructs them back to data space \cite{kingma2014autoencoding}. The LDM operates in this latent space to generate or reconstruct latent sequences conditioned on auxiliary information \cite{rombach2022high}. Finally, the physics corrector, refines the generated fields to improve their physical consistency. Fig. \ref{fig:scheme} presents a schematic view of the proposed framework.

\subsection{Framework structure}

The proposed framework follows a three-stage generative-and-corrective structure. First, a VAE is used to learn a compact representation of the physical field sequences. Second, a conditional LDM generates or reconstructs samples in the learned latent space. Third, a residual corrector operates on the decoded physical fields and refines them using the available conditioning information. This design separates representation learning, generative modeling, and physics-informed refinement into distinct modules, which makes the framework applicable across the different physical tracks considered in this study.

Given an input sequence $x \in \mathbb{R}^{C \times T \times H \times W}$, where \(C\) denotes the number of physical channels, \(T\) the number of time steps, and \(H \times W\) the spatial resolution, the VAE encoder maps the sequence to a lower-dimensional latent representation
$z \in \mathbb{R}^{C_z \times T \times H_z \times W_z}$.
The VAE is implemented as a 3D convolutional encoder--decoder operating on spatiotemporal input sequences, following the standard variational autoencoding framework \cite{kingma2014autoencoding}. The encoder uses repeated 3D convolutions with stride \((1,2,2)\), so that downsampling is applied only in the spatial dimensions while the temporal dimension is preserved. The encoder predicts the parameters of a Gaussian latent distribution, represented by a mean tensor \(\mu\) and a log-variance tensor \(\log \sigma^2\). The decoder mirrors the encoder and progressively upsamples the spatial dimensions to reconstruct the field sequence in the original data space.

The LDM is trained to model the distribution of VAE latents rather than the full-resolution physical fields. Its denoising network is a conditional 3D U-Net operating on noisy latent sequences \cite{cicek20163dunet,rombach2022high,ronneberger2015unet,ho2022videodiffusion}. The U-Net backbone consists of 3D residual blocks with convolutions of kernel size \((1,3,3)\) and padding \((0,1,1)\), which preserves the temporal dimension while extracting spatial features \cite{he2016resnet,cicek20163dunet}. Downsampling and upsampling are likewise applied only over the spatial dimensions. Skip connections between the encoder and decoder paths allow the network to recover fine-scale spatial structure during denoising. Conditioning information is incorporated into the LDM through both vector-valued and spatial inputs. Vector-valued conditioning variables are embedded together with the current noise level and injected into the residual blocks through FiLM modulation \cite{perez2018film}. Spatial conditioning maps are incorporated by concatenation to the U-Net input and by projection into context tokens used in cross-attention layers \cite{rombach2022high}. In addition, spatial and temporal self-attention are applied at selected resolutions, while spatial, temporal, and cross-attention are included in the bottleneck block to support interactions across space, time, and conditioning information \cite{vaswani2017attention,rombach2022high,ho2022videodiffusion}.

After latent generation, the pretrained VAE decoder maps the generated latent sequence back to physical space. The decoded output is then passed to the corrector, which forms the final stage of the framework. The corrected variable is track-specific: for heat conduction it corresponds to flux-related fields, for Navier--Stokes to packed face-velocity and pressure fields, and for the Newtonian gravity setting to scalar gravitational potential fields. Rather than generating a new solution from scratch, the corrector operates in a residual manner. Let \(\mathbf{x}_{\mathrm{pred}}\) denote the decoded LDM prediction. The corrector predicts a residual update \(\Delta \mathbf{x}\), producing the refined output \(\mathbf{x}_{\mathrm{corr}} = \mathbf{x}_{\mathrm{pred}} + \Delta \mathbf{x}\).

Architecturally, the corrector consists of an input projection layer, a sequence of 3D residual blocks, and an output projection layer. In addition to the predicted physical field, the corrector receives the corresponding spatial conditioning maps and vector-valued conditioning variables. These conditioning signals are embedded and concatenated with the input representation, allowing the corrector to adapt its residual refinement to the specific physical parameters and geometry of each sample. The detailed training procedure and the physics-informed objectives used to optimize the corrector are described in the following subsections.

\subsection{Model training}

The proposed framework is trained sequentially in three stages. First, the VAE is trained to learn a compact latent representation of the spatiotemporal field sequences. Second, the LDM is trained on the encoded VAE latents. Third, the physics-informed corrector is trained to refine the decoded diffusion-model outputs in physical space. This staged procedure separates representation learning, generative modeling, and physical refinement, while allowing the same training, validation, and test split to be used consistently across all modules.

For each physical track, a simulated dataset is first generated using the corresponding numerical solver and parameter ranges. The resulting field sequences are partitioned into training, validation, and test subsets. The same split is reused throughout VAE, LDM, and corrector training in order to ensure that the different stages of the pipeline are evaluated on aligned samples.

The VAE is trained on the track-specific physical representation using a \(\beta\)-VAE objective, \(\mathcal{L}_{\mathrm{VAE}}
=
\mathcal{L}_{\mathrm{rec}}
+
\beta \mathcal{L}_{\mathrm{KL}}\),
where \(\mathcal{L}_{\mathrm{rec}}\) is the reconstruction loss and \(\mathcal{L}_{\mathrm{KL}}\) is the Kullback--Leibler divergence between the approximate posterior and a standard Gaussian prior \cite{kingma2014autoencoding,higgins2017beta}. In this work, the reconstruction term is implemented as a mean-squared error over the relevant physical channels, while \(\beta\) is set to a small value in order to prioritize accurate reconstruction while retaining latent regularization. Once trained, the VAE encoder is used to map each physical field sequence to a latent representation, and the corresponding decoder is kept fixed for mapping generated latent samples back to physical space.

Before diffusion training, the encoded latent representations are normalized on a per-channel basis using statistics estimated from the training set. This standardization is used as an implementation step for numerical stability. The diffusion model is then trained using an EDM-style sigma parameterization and preconditioning strategy \cite{karras2022edm}.

The LDM is trained to denoise corrupted VAE latents. Given a clean latent representation \(z_0\), a noisy latent sample is constructed as \(z_{\sigma}
=
z_0 + \sigma \epsilon\) such that \(\epsilon \sim \mathcal{N}(0,I)\), where \(\sigma > 0\) controls the noise magnitude \cite{karras2022edm}. The denoising model is trained primarily to predict the added noise \(\epsilon\), following standard diffusion-model training \cite{ho2020ddpm,karras2022edm}. In addition, an auxiliary reconstruction loss is applied to the estimated clean latent \(\hat{z}_0\). The diffusion training objective is therefore
\(
\mathcal{L}_{\mathrm{LDM}}
=
\lambda_{\epsilon}
\left\|
\hat{\epsilon}
-
\epsilon
\right\|_2^2
+
\lambda_{x_0}
\left\|
\hat{z}_0
-
z_0
\right\|_2^2,
\)
where \(\hat{\epsilon}\) is the predicted noise and \(\hat{z}_0\) is the corresponding estimate of the clean latent. The noise-prediction term provides the main training signal, while the \(x_0\)-loss is used as an auxiliary stabilizing term.

Noise levels are parameterized directly in \(\sigma\)-space and sampled from a log-linearly spaced schedule with \(N=1000\) values between \(\sigma_{\min}=0.02\) and \(\sigma_{\max}=10.0\). This exposes the model to both weakly and strongly corrupted latent samples during training. At inference time, latent sequences are generated by starting from Gaussian noise and iteratively denoising along a decreasing sequence of noise levels. We use an EDM-style Heun sampler with a Karras-type sampling schedule in \(\sigma\)-space \cite{karras2022edm}. The final denoised latent sequence is then decoded using the pretrained VAE decoder.

As seen in algorithm \ref{alg:corrector-training}, after training the VAE and LDM, decoded LDM predictions are generated and cached for the same training, validation, and test samples used in the earlier stages. These cached predictions are paired with the corresponding ground-truth sequences and conditioning variables to train the corrector. The corrector is therefore not trained to generate fields from scratch, but instead to refine the residual errors of the VAE+LDM pipeline in physical space.

The corrector is trained separately from the LDM for two main reasons. First, the two modules have different optimization objectives: the LDM is trained to model the data distribution in latent space, whereas the corrector is optimized to improve physical consistency in data space. Second, the physics-informed losses used by the corrector introduce additional computational and memory costs, making end-to-end joint training impractical. Separating the two stages allows the generative model to be trained independently, after which the corrector can be optimized on cached model outputs.

The corrector objective contains a supervised term that compares the corrected field with the corresponding ground truth. Depending on the physical track, the objective may also include a consistency term that penalizes unnecessary deviations from the original LDM prediction, physics-informed residual terms, and auxiliary regularization terms. In general form, the corrector training objective can be written as
\(
\mathcal{L}_{\mathrm{corr}}
=
w_{\mathrm{sup}}\mathcal{L}_{\mathrm{sup}}
+
w_{\mathrm{phys}}\mathcal{L}_{\mathrm{phys}}
\)
where terms that are not used for a given physical track are omitted. The supervised loss encourages agreement with the ground-truth solution, the consistency term regularizes the magnitude of the correction, the physics-informed loss enforces track-specific physical constraints, and the auxiliary terms are used when additional feature-level or spectral constraints are required. The exact definition of \(\mathcal{L}_{\mathrm{phys}}\) differs between heat conduction, Navier--Stokes, and Newtonian gravity potential, and is described in Section~\ref{sub-sec:PhyInfLoss}.

Training is performed using AdamW optimization, mixed-precision arithmetic when available, and gradient clipping for numerical stability. After each epoch, the model is evaluated on the validation set, and the checkpoint with the best validation performance is retained. The overall corrector training procedure is summarized in Algorithm~\ref{alg:corrector-training}.

\begin{algorithm}
\caption{Training procedure for the physics-informed corrector}
\label{alg:corrector-training}
\begin{algorithmic}[1]
\State Train the VAE on the physical field sequences
\State Encode the training data into VAE latents and estimate latent normalization statistics
\State Train the LDM on normalized latent sequences
\State Decode and cache VAE+LDM predictions for the dataset
\State Initialize corrector parameters \(\theta\)
\For{\(e = 1\) to \(N_{\mathrm{epochs}}\)}
    \For{each mini-batch}
        \State Load decoded prediction \(\mathbf{x}_{\mathrm{pred}}\), ground truth \(\mathbf{x}_{\mathrm{gt}}\), and conditioning variables
        \State Compute corrected field \(\mathbf{x}_{\mathrm{corr}} = f_{\theta}(\mathbf{x}_{\mathrm{pred}}, \mathrm{cond})\)
        \State Compute supervised, physics-informed, and auxiliary loss terms
        \State Form total loss \(\mathcal{L}_{\mathrm{corr}}\)
        \State Backpropagate \(\mathcal{L}_{\mathrm{corr}}\) and update \(\theta\)
    \EndFor
    \State Evaluate the corrector on the validation set
    \If{validation loss improves}
        \State Save current checkpoint
    \EndIf
\EndFor
\end{algorithmic}
\end{algorithm}

Although the three physical tracks share the same overall training procedure, their specific architectures, optimization settings, conditioning variables, and loss terms differ. We therefore report the final per-track settings in Table~\ref{tab:track_model_configs} rather than imposing a single shared configuration across all experiments.

\subsection{Physics-Informed Losses}
\label{sub-sec:PhyInfLoss}
Since the physical constraints differ across the considered datasets, the physics-informed component of the training objective is adapted to each problem class. In the case of heat conduction, the loss function combines supervised and physics-based terms in order to promote both data fidelity and physical admissibility.

\subsubsection{Heat Conduction}
For the heat-conduction setting, the corrector is trained using a combination of supervised, consistency, and physics-informed losses. In addition to the previously stated supervised loss,
$
\mathcal{L}_{\mathrm{sup}},
$
we use the consistency term,
$\mathcal{L}_{\mathrm{stay}}$ to penalize deviations of the corrected flux from the original LDM prediction. This term regularizes the corrector and prevents unnecessary changes to samples that are already close to physically plausible solutions. 

The physics-informed component is composed of several terms tailored to the heat equation. We use $\mathcal{L}_{\mathrm{S}}$ as our smoothing loss responsible for ensuring a smooth transitions between time-steps. This is motivated by the fact that heat conduction produces gradual diffusion of temperature over time, and therefore the corresponding flux field should not contain spurious high-frequency artifacts or discontinuous frame-to-frame changes. The smoothness term helps suppress noisy corrections introduced by the diffusion model or by the corrector itself. The boundary condition loss $\mathcal{L}_{\mathrm{BC}}$ regulates the violation of boundary conditions of the predicted output. This term is important because the boundary behavior strongly influences the evolution of the temperature field, especially under Neumann boundary conditions. Enforcing boundary consistency prevents the corrected flux from implying unphysical heat flow into or out of the domain when the simulated system assumes no such exchange. The PDE residual loss, $\mathcal{L}_{\mathrm{P}}$
penalizes violations of the governing partial differential equation and enforces consistency with the underlying heat-conduction dynamics. This term is motivated by the fact that a flux field may appear plausible locally while still failing to produce a temperature evolution that satisfies the conservation of energy. By penalizing the heat-equation residual, the corrector is encouraged to produce fluxes whose divergence is compatible with the temporal change in temperature. The Fourier loss, $\mathcal{L}_{\mathrm{F}}$
enforces agreement with Fourier's law by comparing the predicted flux to the flux implied by the temperature field. This term directly targets the constitutive structure of heat conduction: heat should flow down the temperature gradient, with magnitude controlled by the thermal conductivity. It therefore prevents the corrected flux from merely matching reference values in a pointwise sense while violating the physical relationship between temperature, conductivity, and heat flow. A temperature reconstruction loss, $\mathcal{L}_{\mathrm{T}}$
measures the discrepancy between the temperature field induced by the corrected flux and the corresponding reference temperature field. This term closes the link between flux correction and the observable thermal evolution. Its purpose is to ensure that the corrected flux is not only physically shaped, but also capable of reconstructing the correct temperature sequence when used to advance or recover the heat dynamics. Finally, the Fourier reconstruction loss, $\mathcal{L}_{\mathrm{FR}}$
computes a flux field from the reconstructed temperature and compares it against the reference flux, thereby encouraging consistency between the corrected flux, the reconstructed temperature, and the constitutive law. This term provides an additional consistency loop between the corrected flux, the reconstructed temperature, and Fourier's law. It encourages the corrected flux to induce a temperature field whose own Fourier-implied flux remains physically consistent with the reference solution.

The overall physics loss is therefore written as
\[
\mathcal{L}_{\mathrm{phys}}
=
w_{\mathrm{P}} \mathcal{L}_{\mathrm{P}}
+
w_{\mathrm{F}} \mathcal{L}_{\mathrm{F}}
+
w_{\mathrm{BC}} \mathcal{L}_{\mathrm{BC}}
+
w_{\mathrm{S}} \mathcal{L}_{\mathrm{S}}
+
w_{\mathrm{T}} \mathcal{L}_{\mathrm{T}}
+
w_{\mathrm{FR}} \mathcal{L}_{\mathrm{FR}},
\]

The complete training objective for heat conduction is thus
\[
\mathcal{L}_{\mathrm{total}}
=
w_{\mathrm{sup}} \mathcal{L}_{\mathrm{sup}}
+
w_{\mathrm{stay}} \mathcal{L}_{\mathrm{stay}}
+
w_{\mathrm{phys}} \mathcal{L}_{\mathrm{phys}}.
\]
This formulation allows the corrector to refine the diffusion-model output while simultaneously enforcing compliance with the governing physics.

\subsubsection{Navier--Stokes}

For the Navier--Stokes setting, the corrector is trained on a packed field representation containing horizontal face velocity, vertical face velocity, and cell-centered pressure. This packed representation is used as the neural-network input and output format for convenience. However, the physics-informed terms are evaluated after unpacking the corrected prediction back to the native staggered marker-and-cell (MAC) layout, since incompressibility, boundary conditions, pressure regularization, and momentum residuals are naturally defined on this grid.

The corrector is trained using a supervised loss, a core physics-informed loss, and an auxiliary loss designed to improve the usefulness of the corrected fields for downstream SR. The supervised term, \(\mathcal{L}_{\mathrm{sup}}\), is implemented as a masked mean-squared error between the corrected field and the corresponding ground-truth field over valid entries of the packed representation. This term anchors the corrected output to the reference simulation and prevents the physics-informed terms from moving the solution toward states that satisfy selected physical constraints but no longer match the simulated data. Unlike the heat-conduction and gravity-potential settings, no explicit consistency loss against the original LDM prediction is used for the reported Navier--Stokes corrector objective.

The core physics-informed component is designed to enforce the main structural properties of incompressible flow. The divergence loss, \(\mathcal{L}_{\mathrm{D}}\), penalizes violations of incompressibility on fluid cells. This term is included because non-zero divergence introduces nonphysical mass imbalance and can degrade the velocity derivatives used later for SR. The momentum loss, \(\mathcal{L}_{\mathrm{M}}\), penalizes residuals of the two-dimensional incompressible Navier--Stokes momentum equations. Although the downstream symbolic-regression analysis focuses on the \(v\)-momentum equation, both horizontal and vertical momentum residuals are used during corrector training. This encourages the corrected velocity field to remain physically coherent as a coupled two-component flow field, rather than only optimizing the component used in the final symbolic-regression experiment. The boundary-condition loss, \(\mathcal{L}_{\mathrm{BC}}\), penalizes deviations from the prescribed inflow profile, non-zero vertical velocity at the inflow, non-zero wall velocities, and non-zero velocities on obstacle-associated faces. This term prevents the corrector from improving the interior flow field at the cost of violating the imposed flow configuration. The pressure-gauge loss, \(\mathcal{L}_{\mathrm{Pg}}\), penalizes non-zero mean pressure over fluid cells, thereby fixing the arbitrary pressure offset inherent in incompressible flow. Finally, the solid-pressure loss, \(\mathcal{L}_{\mathrm{Ps}}\), suppresses pressure values inside solid regions, where pressure is not physically meaningful in the same way as in the fluid domain. Together, these terms stabilize the pressure channel and reduce physically irrelevant pressure artifacts.

The overall Navier--Stokes physics loss is therefore written as
\[
\mathcal{L}_{\mathrm{phys}}
=
w_{\mathrm{D}} \mathcal{L}_{\mathrm{D}}
+
w_{\mathrm{M}} \mathcal{L}_{\mathrm{M}}
+
w_{\mathrm{BC}} \mathcal{L}_{\mathrm{BC}}
+
w_{\mathrm{Pg}} \mathcal{L}_{\mathrm{Pg}}
+
w_{\mathrm{Ps}} \mathcal{L}_{\mathrm{Ps}} .
\]

In addition to the core physics-informed terms, the Navier--Stokes corrector uses auxiliary losses that are specifically motivated by the downstream symbolic-regression task. These losses are grouped into an auxiliary symbolic-regression loss, denoted \(\mathcal{L}_{\mathrm{SR}}\). The purpose of this term is not to replace the physical residuals, but to preserve the derivative structure, term magnitudes, spectral content, and coefficient relationships that SR relies on. In the reported Navier--Stokes experiments, the auxiliary losses are evaluated for the \(v\)-momentum equation using the candidate library
\[
x_0 = -u\frac{\partial v}{\partial x},
\qquad
x_1 = -v\frac{\partial v}{\partial y},
\qquad
x_2 = -\frac{1}{\rho}\frac{\partial p}{\partial y},
\qquad
x_3 = \nabla^2 v .
\]
These terms represent the advective, pressure-gradient, and viscous components from which SR attempts to recover the governing equation. The feature-matching loss, \(\mathcal{L}_{\mathrm{feat}}\), aligns the candidate library terms computed from the corrected field with those computed from the ground-truth field. This term is included because accurate field reconstruction does not necessarily imply accurate derivative-based quantities. Since SR operates on these derived terms, the corrector is explicitly encouraged to preserve the structure of the regression library. The spectral loss, \(\mathcal{L}_{\mathrm{spec}}\), compares the spatial frequency content of the candidate library terms and of the reconstructed right-hand side of the target momentum equation. This term is included to discourage corrections that reduce pointwise error by over-smoothing the field, since such smoothing can remove spatial structures that are important for recovering the governing equation. The amplitude-calibration loss, \(\mathcal{L}_{\mathrm{calib}}\), matches the logarithmic root-mean-square magnitude of selected candidate terms. This is motivated by the fact that SR is sensitive to the relative scale of the library terms. If important terms are systematically amplified or suppressed by the corrector, the downstream regression procedure may assign misleading coefficients or select an incorrect balance of terms. The coefficient-alignment loss, \(\mathcal{L}_{\mathrm{coef}}\), compares the coefficients obtained by fitting the candidate library to the temporal derivative. In practice, a masked ridge-regression problem is solved for both the corrected field and the ground-truth field:
\[
\frac{\partial v}{\partial t}
\approx
c
+
\beta_0 x_0
+
\beta_1 x_1
+
\beta_2 x_2
+
\beta_3 x_3 .
\]
The resulting coefficients are then compared. This loss encourages the corrected data to imply the same equation-level term balance as the reference simulation, which is directly aligned with the downstream symbolic-regression objective. Finally, the correlation-alignment loss, \(\mathcal{L}_{\mathrm{corr}}\), compares the masked Pearson correlation between each candidate library term and the temporal derivative \(\partial v/\partial t\). This term preserves the statistical relationship between the candidate terms and the observed temporal evolution. It complements the coefficient-alignment loss by matching the explanatory structure of the regression library, rather than only the fitted coefficient values. 

The auxiliary symbolic-regression loss is therefore written as
\[
\mathcal{L}_{\mathrm{SR}}
=
w_{\mathrm{feat}} \mathcal{L}_{\mathrm{feat}}
+
w_{\mathrm{spec}} \mathcal{L}_{\mathrm{spec}}
+
w_{\mathrm{calib}} \mathcal{L}_{\mathrm{calib}}
+
w_{\mathrm{coef}} \mathcal{L}_{\mathrm{coef}}
+
w_{\mathrm{corr}} \mathcal{L}_{\mathrm{corr}} .
\]

The complete Navier--Stokes corrector objective is thus
\[
\mathcal{L}_{\mathrm{total}}^{\mathrm{NS}}
=
w_{\mathrm{sup}} \mathcal{L}_{\mathrm{sup}}
+
w_{\mathrm{phys}} \mathcal{L}_{\mathrm{phys}}
+
\mathcal{L}_{\mathrm{SR}} .
\]
This formulation allows the corrector to refine the diffusion-model output while preserving three properties required for the downstream task: agreement with the reference simulation, consistency with the incompressible Navier--Stokes structure, and preservation of the derivative-based quantities used for SR.

\subsubsection{Single-Mass Newtonian Gravitational Potential}
For the single-mass Newtonian-gravity setting, the corrector is trained using a combination of supervised, consistency, and physics-informed losses applied directly in the space of scalar potential fields. In addition to the previously stated supervised loss, $\mathcal{L}_{\mathrm{sup}},$ we use the consistency term, $\mathcal{L}_{\mathrm{stay}},$ which penalizes deviations of the corrected potential from the original LDM prediction. This term regularizes the corrector and discourages unnecessary modifications to samples that are already close to plausible solutions.

The physics-informed component is tailored to the fact that the target quantity is the scalar gravitational potential \(\Phi(x,y,t)\), rather than a vector-valued field. We therefore do not enforce a constitutive relation such as Fourier's law. Instead, we impose consistency with the analytic single-mass potential model and with the corresponding inferred source trajectory. First, we use a smooth inverse-fit procedure to estimate the source trajectory from the corrected potential sequence. Concretely, source positions are extracted from each frame using a softargmin over the potential field, after which a constant-velocity trajectory is fitted across time. Given this fitted trajectory, and either the known conditioning mass or an estimated mass, an analytic potential sequence is rendered according to the single-mass Newtonian potential law.

The physics-informed loss is composed of several terms. We use \(\mathcal{L}_{\mathrm{S}}\) as a smoothness loss that penalizes abrupt temporal and spatial variations in the corrected potential field.  This is motivated by the fact that, away from the source singularity or softened source region, the Newtonian potential should vary smoothly in space and evolve smoothly over time as the mass moves. The smoothness term therefore suppresses high-frequency artifacts that may be introduced by the diffusion model or by the corrector itself. The boundary-condition loss \(\mathcal{L}_{\mathrm{BC}}\) penalizes discrepancies between the corrected potential and the analytic reference potential along the spatial boundaries of the domain. This term is included because boundary regions provide a useful global constraint on the field. Although the source is localized, the gravitational potential is nonlocal, and errors near the boundary can indicate that the corrected field has the wrong large-scale amplitude, offset, or spatial decay. Enforcing boundary consistency therefore helps prevent corrections that match the source region locally while producing an implausible global potential field. The Poisson-structure loss, \(\mathcal{L}_{\mathrm{P}}\), compares the discrete Laplacian of the corrected potential against the Laplacian of the reference analytic potential and thereby encourages consistency with the underlying source-induced field structure. This loss is included because of the fact that the gravitational potential is not an arbitrary scalar image, rather its curvature is tied to the source distribution. The analytic-law loss, \(\mathcal{L}_{\mathrm{L}}\), penalizes discrepancies between the corrected potential and the potential reconstructed from the inverse-fitted single-mass trajectory, thus encouraging the corrected output itself to remain explainable by the intended physical law. This term directly enforces explainability by the intended Newtonian potential law. In other words, the corrected sequence should not simply resemble the ground-truth field pointwise; it should be representable as the potential generated by a single moving mass. This helps rule out corrections that reduce pixel-wise error while introducing field patterns that cannot be attributed to any valid single-source trajectory.

In addition, we introduce a trajectory-reconstruction loss, $
\mathcal{L}_{\mathrm{TR}},$
which compares the inverse-fitted source trajectory against the ground-truth trajectory. This term includes both source positions over time and the fitted constant velocity, thereby encouraging the corrected potential sequence to imply the correct motion of the source mass. This encourages the network to not introduce additional forces into the environment which would cause the acceleration of the mass. Finally, a potential-reconstruction loss,
 $
\mathcal{L}_{\Phi\mathrm{R}},$
measures the discrepancy between the analytically reconstructed potential induced by the fitted trajectory and the corresponding ground-truth potential sequence. This encourages consistency between the corrected field, the inferred latent motion, and the reference physical evolution. This term closes the loop between field correction, inverse trajectory estimation, and forward physical rendering. It ensures that the trajectory inferred from the corrected field, when passed back through the analytic Newtonian potential model, reproduces the correct physical evolution. This encourages consistency between the corrected potential field, the inferred latent motion, and the reference dynamics.

The overall physics loss is therefore written as
\[
\mathcal{L}_{\mathrm{phys}}
=
w_{\mathrm{P}} \mathcal{L}_{\mathrm{P}}
+
w_{\mathrm{L}} \mathcal{L}_{\mathrm{L}}
+
w_{\mathrm{BC}} \mathcal{L}_{\mathrm{BC}}
+
w_{\mathrm{S}} \mathcal{L}_{\mathrm{S}}
+
w_{\mathrm{TR}} \mathcal{L}_{\mathrm{TR}}
+
w_{\Phi\mathrm{R}} \mathcal{L}_{\Phi\mathrm{R}}.
\]

The complete training objective for the single-mass gravitational-potential corrector is thus
\[
\mathcal{L}_{\mathrm{total}}
=
w_{\mathrm{sup}} \mathcal{L}_{\mathrm{sup}}
+
w_{\mathrm{stay}} \mathcal{L}_{\mathrm{stay}}
+
w_{\mathrm{phys}} \mathcal{L}_{\mathrm{phys}}.
\]

This formulation allows the corrector to refine the diffusion-model output while simultaneously enforcing compliance with the structure of the scalar Newtonian potential field and with the source trajectory that generated it.

\section{Experimental Design}
\label{sec:experiments}
In order to evaluate the proposed framework, we consider three experiments: (i) equation discovery under temporal sparsity using synthetic data enrichment for SR, (ii) similarly equation recovery from purely synthetic sequences and (iii) assessment of the physical validity and fidelity of the generative models and physics-informed correctors. These experiments assess whether the proposed models can meaningfully complete under-observed trajectories, which is the necessary but not sufficient condition for data enrichment for SR. In addition, it evaluates the proposed framework ability to produce synthetic data that is useful for downstream SR tasks as well as preserve the physical structure of the underlying dynamical system rather than merely reproducing statistically plausible patterns. Moreover, aiming to obtain a robust evaluation, we conduct each of the experiments on three physical dynamics: heat conduction, incompressible Navier-Stokes flow, and single-mass Newtonian gravitational potential. These systems were selected to provide a diverse range of physical structures and modeling difficulty, spanning constitutive-law recovery, PDE-term recovery, and recovery of a scalar potential law induced by a moving source. Furthermore, we used three popular and out-of-the-box SR models: GPLearn, DEAP, and PySR.

\begin{figure}
    \centering
    \includegraphics[width=0.99\linewidth]{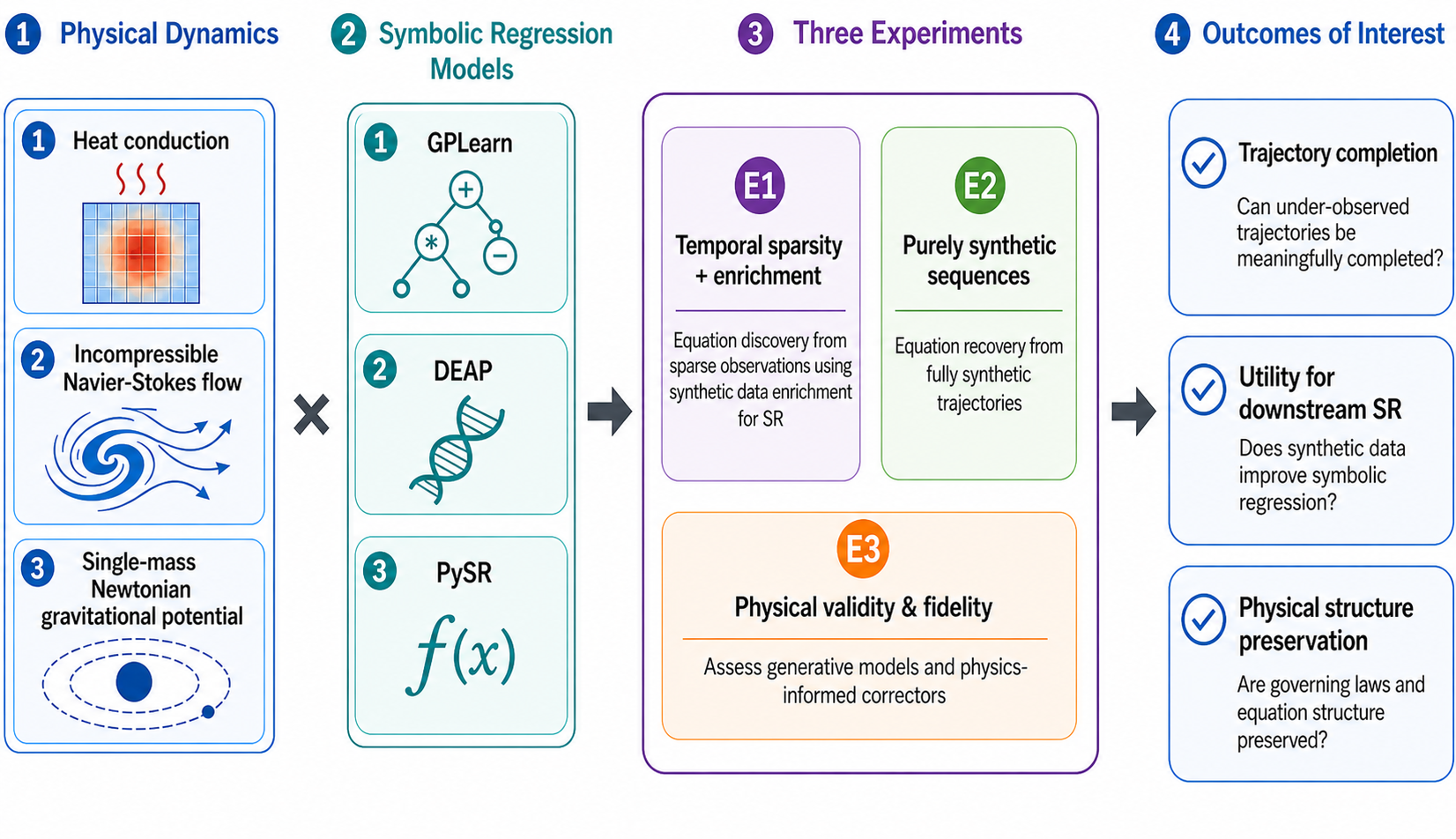}
    \caption{A schematic view of the experimental flow. }
    \label{fig:scheme_exp}
\end{figure}

Below, we first formally describe the data generation processes of the three physical experiments, followed by a description of the four experiments. Fig. \ref{fig:scheme_exp} presents a schematic view of the experimental flow. 

\subsection{Physical dynamics}
The proposed framework is evaluated on three physical dynamics: heat conduction, incompressible Navier--Stokes flow, and single-mass Newtonian gravitational potential. These physical dynamics were selected to cover different forms of physical structure and SR difficulty. Heat conduction evaluates recovery of a constitutive law, Navier--Stokes evaluates recovery of PDE terms in a coupled velocity--pressure system, and the gravitational-potential setting evaluates recovery of a scalar potential law induced by a moving source.

Although the framework is shared across all three physical tracks, the exact model sizes, optimization settings, data representations, and loss terms differ between systems. Table~\ref{tab:track_model_configs} summarizes the core architecture and training configurations used for the three-stage generative pipeline. Track-specific deviations from the general training setup, such as the derivative-aware Navier--Stokes VAE objective and the longer gravitational-potential LDM training, are described below.

\begin{table*}[!ht]
\centering
\caption{Experimental configurations for the proposed framework used for each physical dynamics.}
\label{tab:track_model_configs}
\small
\begin{adjustbox}{max width=\textwidth}
\begin{tabular}{ccp{9.7cm}}
\hline \hline
\textbf{Physical dynamics} & \textbf{Module} & \textbf{Configuration} \\
\hline \hline
 & VAE &
epochs \(=100\), learning rate \(=1\cdot 10^{-4}\), batch size \(=64\), base channels \(=32\), latent dimension \(=64\) \\

Heat conduction & LDM &
epochs \(=150\), learning rate \(=1\cdot 10^{-4}\), batch size \(=8\), latent diffusion in VAE latent space \\

 & Corrector &
epochs \(=100\), learning rate \(=1\cdot 10^{-4}\), batch size \(=2\), base channels \(=64\), depth \(=2\) \\
\hline
 & VAE &
epochs \(=140\), learning rate \(=8\cdot 10^{-5}\), batch size \(=24\), base channels \(=48\), latent dimension \(=128\), spatial downsampling stages \(=2\) \\

Navier--Stokes & LDM &
epochs \(=180\), learning rate \(=1.2\cdot 10^{-4}\), batch size \(=8\), latent diffusion in VAE latent space, \(x_0\)-auxiliary weight \(=0.9\) \\

 & Corrector &
epochs \(=160\), learning rate \(=1.4\cdot 10^{-4}\), weight decay \(=5\cdot 10^{-5}\), batch size \(=16\), base channels \(=32\), depth \(=4\) \\
\hline
 & VAE &
epochs \(=80\), learning rate \(=1\cdot 10^{-4}\), batch size \(=8\), base channels \(=64\), latent dimension \(=64\) \\

Gravitational potential & LDM &
epochs \(=500\), learning rate \(=2\cdot 10^{-4}\), batch size \(=8\), latent diffusion in VAE latent space \\

 & Corrector &
epochs \(=100\), learning rate \(=2\cdot 10^{-4}\), batch size \(=8\), depth \(=6\) \\
\hline \hline
\end{tabular}
\end{adjustbox}
\end{table*}

\subsubsection{Heat Conduction}
\label{sec:exp_heat}
For the heat-conduction physical dynamics, we generated synthetic spatiotemporal sequences by numerically solving a two-dimensional heat-conduction system on a \(128 \times 128\) grid using an explicit finite-difference scheme with insulated Neumann boundary conditions. Each sequence consists of 32 saved time steps. Internally, the solver advances the system with a smaller stable time step and stores every \(k\)-th state, where \(k=128\) in the dataset-generation pipeline used for training the generative models and for the SR experiments reported here.

The temperature field was initialized as spatially uniform, with the initial temperature sampled independently for each sequence from the interval \([0,50]\). Thermal conductivity was sampled as a spatially heterogeneous piecewise-constant field. Specifically, a base conductivity was first sampled from the interval \([5,50]\), after which the domain was partitioned into a \(2 \times 2\) patch grid and each patch was multiplied by an independently sampled random factor. The resulting conductivity field was clipped to the interval \([5,50]\), which also guaranteed stability under the globally chosen explicit time step.

Heating was induced by a localized source represented as a uniform disk with random radius and random center location. The source radius was sampled from \(30\) to \(50\) cells, and the total source power was sampled from the interval \([500,1000]\). The disk mask was normalized over its support so that the sampled source power corresponds to the total injected power per unit time rather than a per-cell intensity. The source remained active throughout the full sequence with a constant temporal profile.

The global simulation time step was chosen according to a worst-case stability rule based on the maximum allowed conductivity of \(\Delta t = 0.8\Delta x^2/8k_{\max}\), such that \(k_{\max}=50\). This ensured numerical stability for all generated sequences under a common discrete time step. For each saved frame, we stored the temperature field \(T\), its first- and second-order temporal finite differences, and the corresponding heat fluxes. In addition to cell-centered fluxes, we also stored the face fluxes \((q_x^+, q_x^-, q_y^+, q_y^-)\), which are later used to reconstruct temperature fields consistently from predicted synthetic flux sequences.

For conditional generation, each sequence is represented not only by its spatiotemporal flux observations but also by conditioning variables derived from the underlying physical configuration. These include the conductivity field and transformations thereof, the source mask and its geometry, and scalar descriptors such as source power, source radius, conductivity statistics, source area fraction, and source location. Both the LDM and the physics-informed corrector use these conditioning variables when generating sequence completions.

For the SR, the objective is to recover Fourier's law from spatiotemporal observations. In particular, we seek to recover the constitutive relations \(q_x = -k \frac{\partial T}{\partial x}\) and \(q_y = -k \frac{\partial T}{\partial y}\), where \(T\) denotes temperature, \(k\) denotes thermal conductivity, and \(q_x\) and \(q_y\) denote the horizontal and vertical heat-flux components, respectively.

In order to construct the SR samples, spatial temperature gradients are computed from each frame and paired with the conductivity field. For the \(q_x\) equation, the regression inputs are features derived from \(\partial T/\partial x\) and \(k\), with \(q_x\) as the target. For the \(q_y\) equation, the regression inputs are features derived from \(\partial T/\partial y\) and \(k\), with \(q_y\) as the target. For synthetic top-up variants, temperature fields are reconstructed from the predicted face-flux trajectories using the same discrete update rule as in the forward simulator, ensuring that the SR inputs remain consistent with the predicted flux sequence and the governing heat dynamics.

\subsubsection{Navier-Stokes}
\label{sec:exp_ns}
For the Navier-Stokes physical dynamics, we generated synthetic spatiotemporal sequences by numerically solving the two-dimensional incompressible Navier-Stokes equations on a staggered MAC grid \cite{harlow1965mac}. The computational grid size was \(128 \times 128\), and the physical domain had extent \(L_x=4.0\) and \(L_y=1.0\). Simulations were performed with fixed density \(\rho=1.0\) and time step \(\Delta t = 2.0\cdot 10^{-3}\). A projection-based solver was used, combining semi-Lagrangian advection, explicit diffusion, and pressure projection to enforce incompressibility.

The primary data-generation regime corresponds to moderate-Reynolds-number flow past a circular obstacle. For each sequence, the maximum inflow velocity was sampled from \([0.55,0.90]\), while the obstacle radius was sampled from \([0.075,0.105]\). The obstacle center was fixed at \((x_c,y_c)=(0.5L_x,0.5)\), and the inflow profile was chosen to be Poiseuille. Reynolds numbers were sampled independently from \([80,250]\), after which the viscosity was determined from the sampled inflow velocity and obstacle diameter according to \(\nu = \frac{U_{\max}(2r)}{\mathrm{Re}}\) such that \(\mu = \rho \nu\). Due to the fact that we set \(\rho=1.0\) in all simulations, this yields a variable-viscosity regime parameterized directly by the sampled Reynolds number.

No-slip boundary conditions were enforced at the upper and lower walls and on obstacle-adjacent faces. At the inflow boundary, the horizontal velocity was prescribed by the Poiseuille profile and the vertical velocity was fixed to zero. At the outflow boundary, the corresponding pressure condition was imposed through the projection step. No external body force was applied in the main regime.

Each rollout was generated for 32 coarse temporal intervals. Internally, the solver stores the initial state and all subsequent coarse observations, resulting in \(T+1\) saved states per rollout. In the learning pipeline, the first \(T=32\) saved frames are used as the training sequence. For each saved frame, we stored the cell-centered horizontal velocity \(u\), vertical velocity \(v\), and pressure \(p\), together with the native MAC face velocities. We also stored the obstacle mask, a signed-distance field describing the obstacle geometry, and simulation metadata including \(\Delta x\), \(\Delta y\), \(\Delta t\), inflow velocity, Reynolds number, viscosity, and obstacle size.

Although the simulator produces cell-centered velocity-pressure observations, the active representation used for the generative models was based on staggered face velocities and pressure. More specifically, each frame was represented by three fields: the horizontal face velocity \(u_{\mathrm{face}}\), the vertical face velocity \(v_{\mathrm{face}}\), and the cell-centered pressure field \(p\). These fields were embedded into a common canvas representation so that all channels share the same tensor shape during training.

Because the MAC face fields have native shapes \((H,W+1)\) and \((H+1,W)\), whereas the pressure field is defined on \((H,W)\), the three fields were packed into a shared \(160 \times 160\) canvas. This representation preserves the staggered structure of the velocity field while allowing the VAE and LDM to operate on a uniform tensor layout. Before packing, the pressure field was centered per frame by subtracting its mean over fluid cells, and obstacle cells were set to zero. The resulting learned tensor therefore has shape \( (T,3,H_{\mathrm{pack}},W_{\mathrm{pack}})\), with \(T=32\) and \((H_{\mathrm{pack}},W_{\mathrm{pack}})=(160,160)\).

For conditional generation, each Navier--Stokes sequence was paired with both spatial conditioning maps and a low-dimensional conditioning vector. The spatial conditioning maps consisted of the obstacle mask, the obstacle signed-distance field, the inflow profile map, and relative coordinate maps \(x_{\mathrm{rel}}\) and \(y_{\mathrm{rel}}\) measured with respect to the obstacle center. Together, these maps encode domain geometry, boundary forcing, and obstacle-relative spatial position. In addition, each sequence was associated with a conditioning vector containing normalized scalar descriptors of the physical regime, including the kinematic viscosity, the maximum inflow velocity, and an obstacle-size descriptor derived from the sampled radius.

For this physical dynamics, the SR experiments were performed on the vertical-momentum equation only. The target equation is
\begin{equation}
\frac{\partial v}{\partial t}
=
- u\frac{\partial v}{\partial x}
- v\frac{\partial v}{\partial y}
- \frac{1}{\rho}\frac{\partial p}{\partial y}
+ \nu \nabla^2 v.
\end{equation}

To construct SR rows, the generated or ground-truth fields are first converted into the representation used for derivative estimation. Temporal derivatives are estimated directly from the sequence, while spatial terms are computed from the velocity and pressure fields. The candidate feature library is
\[
\phi_v
=
\left\{
-u\frac{\partial v}{\partial x},
\;
-v\frac{\partial v}{\partial y},
\;
-\frac{1}{\rho}\frac{\partial p}{\partial y},
\;
\nabla^2 v
\right\},
\]
with target \(\partial_t v\). 

\subsubsection{Single-Mass Newtonian Gravitational Potential}
\label{sec:exp_gravity}
For the single-mass Newtonian-gravity physical dynamics, we generated synthetic spatiotemporal sequences by evaluating the time-dependent Newtonian gravitational potential induced by a single moving mass on a two-dimensional \(128 \times 128\) grid. Each sequence consists of 32 saved time steps. In contrast to the previous tracks, the target quantity here is not a vector-valued field but a scalar field, namely the gravitational potential \(\Phi(x,y,t)\). The corresponding gravitational field would be obtained from the spatial gradient of this potential; however, in the present track the generative models and SR experiments operate directly on the scalar potential field.

For each sequence, the source mass was sampled from the interval \([20,80]\), and a softening parameter was sampled from \([3.5,7.0]\) in order to regularize the singularity at the source location and keep the field numerically well behaved in the vicinity of the moving mass. The saved time step \(\Delta t_{\mathrm{saved}}\) was sampled independently from \([0.90,1.10]\). The source was constrained to follow a constant-velocity trajectory that remained entirely within the simulation domain over the full 32-step sequence. To achieve this, a motion direction was first sampled, after which a feasible speed \(v\) was drawn from \([2.5,4.5]\) subject to the requirement that the full trajectory remain inside a box excluding a boundary margin of 10 cells on all sides. An initial position was then sampled so that both the starting point and the final point satisfied this interior constraint.

Given the sampled parameters, the source trajectory is defined by \(x_m(t) = x_0 + v_x t\) and \(y_m(t) = y_0 + v_y t\), and the scalar potential field is generated as \(\Phi(x,y,t) := -Gm/(\sqrt{(x-x_m(t))^2 + (y-y_m(t))^2 + \varepsilon^2})\), where \(m\) denotes the source mass, \(G\) is the gravitational constant used in the simulator, and \(\varepsilon\) denotes the softening parameter. In all experiments reported here, \(G\) was fixed to \(1.0\). For each generated sequence, we stored the full potential trajectory together with the underlying source trajectory, including positions, velocities, and accelerations, as well as the sampled physical parameters such as mass, softening, initial position, final position, and saved time step.

For conditional generation, each sequence is represented not only by its spatiotemporal potential observations but also by conditioning variables derived from the underlying physical configuration. These include trajectory-related conditioning signals describing the source motion, together with scalar descriptors such as mass, softening, and \(\Delta t_{\mathrm{saved}}\). Both the LDM and the physics-informed corrector use these conditioning variables when generating sequence completions.

Directly applying SR to \(\Phi(x,y,t)\) would encourage the algorithms to absorb the product \(Gm\) into a fitted numerical constant, rather than recover the underlying geometric structure of the potential field. To focus the regression problem on the physically meaningful radial dependence, we define the recovery target as the inverse-scaled potential:
\begin{equation}
-\frac{Gm}{\Phi(x,y,t)}
=
\sqrt{(x-x_m(t))^2 + (y-y_m(t))^2 + \varepsilon^2}.    
\end{equation}
This transformation removes the constant amplitude term and makes the SR task equivalent to recovering the softened source-relative distance. As a result, the SR models are evaluated on whether they recover the correct spatial structure of the Newtonian potential, rather than on whether they can fit a sequence-specific multiplicative constant.

\subsection{Numerical analysis}

Our primary objective is to evaluate whether, and to what extent, synthetic data enrichment improves downstream SR when only a small number of real observations are available. To this end, we simulate temporal sparsity by retaining only \(N \in \{2,4,8,16\}\) real frames from each sequence and completing the remaining frames with synthetic data until the original sequence length is restored. For each source sequence, we construct three paired variants: (i) \emph{Real only}, in which SR is performed using only the retained real frames; (ii) \emph{Real+LDM}, in which the missing frames are filled with samples generated by the LDM; and (iii) \emph{Real+PhysLDM}, in which the missing frames are first generated by the LDM and subsequently refined by the physics-informed corrector. In all cases, the synthetic sequences are generated using the same conditioning inputs as the corresponding ground-truth sequence, so that the completion remains matched to the same underlying physical scenario.

All SR models are trained separately on each variant and evaluated against the full ground-truth version of the corresponding source sequence. This gives a paired experimental design in which each \((\text{source sequence}, \text{seed})\) pair defines one experimental unit. Results are aggregated over \(n=30\) source sequences and \(c=5\) random seeds, resulting in 150 paired runs for each sparsity level and SR model.

Across the SR experiments, the recovered symbolic expression is evaluated using mean squared error (MSE) and coefficient of determination \(R^2\):
\(
\mathrm{MSE}
=
\frac{1}{M}
\sum_{i=1}^{M}
\left(
\hat{y}_i-y_i
\right)^2
\) and \(
R^2
=
1
- \sum_{i=1}^{M}
\left(
\hat{y}_i-y_i
\right)^2 / \sum_{i=1}^{M}
\left(
y_i-\bar{y}
\right)^2
\), where \(y_i\) is the ground-truth target value, \(\hat{y}_i\) is the value predicted by the recovered symbolic expression, and \(M\) is the number of evaluation points. The interpretation of \(y\) is track-specific. For heat conduction, the target is the heat-flux relation used to evaluate recovery of Fourier's law, and the reported metrics are denoted \(q\)-MSE and \(q\)-\(R^2\). For Navier--Stokes, the target is the temporal derivative \(\partial v/\partial t\) in the vertical-momentum equation, and the reported metrics are denoted \(\partial_t v\)-MSE and \(\partial_t v\)-\(R^2\). For the gravitational-potential setting, the target is the transformed scalar-potential law introduced in Section~\ref{sec:exp_gravity}, and the reported metrics are denoted \(\phi\)-MSE and \(\phi\)-\(R^2\).

In addition to the temporal-sparsity experiment, we evaluate whether the SR models can recover the underlying equations from purely synthetic sequences. For each source sequence, we construct two variants: \emph{LDM}, in which the full sequence is generated by the LDM, and \emph{PhysLDM}, in which the generated sequence is subsequently refined by the physics-informed corrector. This experiment uses the same track-specific MSE and \(R^2\) metrics as the temporal-sparsity experiment, but removes retained real frames entirely.

Finally, we evaluate the physical validity and fidelity of both the LDM and the PhysLDM. Since the three physical dynamics involve different physically meaningful quantities, this analysis uses track-specific metrics. For heat conduction, we report field-level flux and reconstructed-temperature fidelity, together with flux-divergence and Fourier-law residual metrics. For Navier--Stokes, we report packed-field fidelity and physical-validity metrics related to incompressibility, momentum residuals, boundary conditions, pressure gauge consistency, pressure values inside solid regions, and solid-face velocities. For the single-mass Newtonian-potential setting, we report trajectory, velocity, spectral, and global physical-quantity errors. These metrics are used to evaluate whether the generated samples preserve the physical structure required by the downstream SR task.

Table~\ref{tab:sr_hparams} summarizes the SR hyperparameters used for the experiments.

\begin{table}[!ht]
\centering
\caption{Hyperparameters used for the symbolic-regression experiments.}
\label{tab:sr_hparams}
\small
\begin{adjustbox}{max width=\textwidth}
\begin{tabular}{ccp{9.3cm}}
\hline \hline
\textbf{Physical dynamics} & \textbf{SR model} & \textbf{Hyperparameters} \\
\hline \hline
 & DEAP & population size \(=300\), generations \(=12\), tournament size \(=7\), maximum tree height \(=6\), complexity penalty \(=10^{-4}\) \\
Heat conduction & GPLearn & population size \(=500\), generations \(=8\) \\
 & PySR & iterations \(=3\), population size \(=20\), parsimony \(=0.0\), maximum size \(=30\), model selection \(=\textit{best}\) \\
\midrule
 & DEAP & population size \(=300\), generations \(=12\), tournament size \(=7\), initial tree depth \(=1\text{--}4\), maximum tree height \(=6\), constants in \([-0.25,0.25]\), crossover probability \(=0.7\), mutation probability \(=0.2\), complexity penalty \(=10^{-4}\), function set \(=\{\texttt{add},\texttt{sub},\texttt{mul},\texttt{div}\}\) \\
Navier--Stokes & GPLearn & population size \(=500\), generations \(=8\), tournament size \(=20\), parsimony coefficient \(=0.001\), constants in \([-0.25,0.25]\), initial tree depth \(=1\text{--}4\) \\
 & PySR & iterations \(=3\), population size \(=20\), parsimony \(=0.0\), maximum size \(=30\), model selection \(=\textit{best}\) \\
\midrule
 & DEAP & population size \(=300\), generations \(=12\), tournament size \(=7\), initial tree depth \(=1\text{--}4\), maximum tree height \(=6\), constants in \([-0.25,0.25]\), crossover probability \(=0.7\), mutation probability \(=0.2\), complexity penalty \(=10^{-4}\), function set \(=\{\texttt{add},\texttt{sub},\texttt{mul},\texttt{div}\}\) \\
Gravitational potential & GPLearn & population size \(=500\), generations \(=8\), tournament size \(=20\), parsimony coefficient \(=0.001\), constants in \([-0.25,0.25]\), initial tree depth \(=1\text{--}4\) \\
 & PySR & iterations \(=3\), population size \(=20\), parsimony \(=0.0\), maximum size \(=30\), model selection \(=\textit{best}\) \\
\hline \hline
\end{tabular}
\end{adjustbox}
\end{table}

\section{Results}
\label{sec:results}
Table~\ref{tab:sparsity_equation_discovery} summarizes equation-discovery performance under temporal sparsity across the three physical dynamics. For heat conduction, performance is reported using flux-prediction error and \(q\)-\(R^2\). For Navier--Stokes, the metrics compare the output of the recovered symbolic expression with the target temporal derivative \(\partial v/\partial t\) for the vertical-momentum equation, evaluated on the full ground-truth sequence. For the gravitational-potential task, the metrics quantify recovery of the transformed scalar potential law. One can notice several patterns across the three physical dynamics. For heat conduction, Real+PhysLDM is consistently best across all metrics, SR models, and sparsity levels. The improvement is particularly noticeable in the low-observation regime, where Real-only has insufficient temporal coverage, and Real+LDM fails to recover Fourier's law. The corrected samples reduce \(q\)-MSE to approximately \(10^{-4}\) and raise \(q\)-\(R^2\) close to one. For Navier--Stokes, the benefit of physics-corrected completion is concentrated in the highly sparse regimes. At \(N=2\) and \(N=4\), Real+PhysLDM gives the best \(\partial_t v\)-MSE and \(\partial_t v\)-\(R^2\) for all three SR models. At \(N=8\) and \(N=16\), Real-only becomes the strongest variant, suggesting that once enough true frames are available, the remaining synthetic completion is less beneficial for recovering the vertical-momentum equation. Across all sparsity levels, Real+LDM performs poorly, often producing negative \(R^2\), indicating poor agreement between the recovered symbolic expression and the target temporal derivative. For the gravitational-potential task, Real+PhysLDM generally gives the best performance for GPLearn and PySR across all sparsity levels. For DEAP, Real-only slightly outperforms Real+PhysLDM at \(N=4\), \(N=8\), and \(N=16\), although Real+PhysLDM remains substantially better than Real+LDM. This indicates that physics correction improves the generated potential fields, but the effect on SR also depends on the SR model's sensitivity to the generated samples and on the amount of real data already available.

\begin{table*}[t]
\centering
\caption{Sparsity-dependent equation discovery performance across physical dynamics, symbolic regression backends, and data-enrichment variants.}
\label{tab:sparsity_equation_discovery}
\scriptsize
\setlength{\tabcolsep}{3pt}
\renewcommand{\arraystretch}{1.08}

\begin{adjustbox}{max width=\textwidth}
\begin{tabular}{lll c ccc ccc ccc}
\toprule
\multirow{2}{*}{\textbf{Physical dynamics}} &
\multirow{2}{*}{\textbf{Metric}} &
\multirow{2}{*}{\textbf{Measurement}} &
\multirow{2}{*}{\(\boldsymbol{N}\)} &
\multicolumn{3}{c}{\textbf{GPLearn}} &
\multicolumn{3}{c}{\textbf{DEAP}} &
\multicolumn{3}{c}{\textbf{PySR}} \\
\cmidrule(lr){5-7}
\cmidrule(lr){8-10}
\cmidrule(lr){11-13}
& & & &
Real only & Real+LDM & Real+PhysLDM &
Real only & Real+LDM & Real+PhysLDM &
Real only & Real+LDM & Real+PhysLDM \\
\midrule

\multirow{8}{*}{Heat conduction}
& MSE &\(q\)  & 2  & 0.1266 & 0.1285 & \textbf{0.000213} & 0.1014 & 0.1335 & \textbf{0.000211} & 0.0922 & 0.1041 & \textbf{0.000216} \\
&  MSE &\(q\)  & 4  & 0.0204 & 0.1277 & \textbf{0.000213} & 0.00847 & 0.1329 & \textbf{0.000197} & 0.0276 & 0.1045 & \textbf{0.000245} \\
&  MSE &\(q\)  & 8  & 0.0005 & 0.1266 & \textbf{0.000203} & 0.00196 & 0.1285 & \textbf{0.000140} & 0.00449 & 0.1016 & \textbf{0.000202} \\
&  MSE &\(q\)  & 16 & 0.00024 & 0.1190 & \textbf{0.000203} & 0.000146 & 0.1123 & \textbf{0.0000847} & 0.000242 & 0.0935 & \textbf{0.0000915} \\
\addlinespace

&  \(R^2\) &\(q\) & 2  & -0.077 & -0.1542 & \textbf{0.9667} & 0.1954 & -0.1191 & \textbf{0.9987} & 0.2524 & 0.0999 & \textbf{0.9981} \\
& \(R^2\) &\(q\) & 4  & 0.7141 & -0.1467 & \textbf{0.9667} & 0.9063 & -0.1100 & \textbf{0.9987} & 0.7850 & 0.1038 & \textbf{0.9980} \\
& \(R^2\) &\(q\) & 8  & 0.9632 & -0.1404 & \textbf{0.9670} & 0.9865 & -0.0749 & \textbf{0.9991} & 0.9657 & 0.1283 & \textbf{0.9988} \\
& \(R^2\) &\(q\)  & 16 & 0.9656 & -0.0981 & \textbf{0.9670} & 0.9991 & 0.0977 & \textbf{0.9994} & 0.9978 & 0.2115 & \textbf{0.9994} \\
\addlinespace

\midrule

\multirow{8}{*}{Navier--Stokes}
&  MSE & \(\partial_t v\)  & 2  & 0.4160 & 2.0087 & \textbf{0.1741} & 0.4064 & 6.1498 & \textbf{0.1371} & 0.4060 & 1.6032 & \textbf{0.1894} \\
&   MSE & \(\partial_t v\)   & 4  & 0.2380 & 87.9569 & \textbf{0.1691} & 0.1935 & 2.4223 & \textbf{0.1368} & 0.2614 & 1.4997 & \textbf{0.1877} \\
&   MSE & \(\partial_t v\)   & 8  & \textbf{0.1061} & 1.5939 & 0.1618 & \textbf{0.1127} & 1.9494 & 0.1358 & \textbf{0.1009} & 1.3784 & 0.1827 \\
&   MSE & \(\partial_t v\)   & 16 & \textbf{0.0889} & 1.5085 & 0.2060 & \textbf{0.0961} & 1.9381 & 0.1424 & \textbf{0.0799} & 1.5692 & 0.1927 \\
\addlinespace

& \(R^2\) &\(\partial_t v\)   & 2  & 0.2259 & -2.5009 & \textbf{0.6859} & 0.2197 & -10.7547 & \textbf{0.7413} & 0.2195 & -2.0730 & \textbf{0.6310} \\
& \(R^2\) &\(\partial_t v\)   & 4  & 0.5630 & -175.4489 & \textbf{0.6989} & 0.6333 & -3.5675 & \textbf{0.7410} & 0.5100 & -1.8672 & \textbf{0.6359} \\
& \(R^2\) &\(\partial_t v\)   & 8  & \textbf{0.8053} & -1.7161 & 0.7109 & \textbf{0.7875} & -2.7928 & 0.7445 & \textbf{0.8117} & -1.6756 & 0.6463 \\
& \(R^2\) &\(\partial_t v\)   & 16 & \textbf{0.8371} & -1.6803 & 0.6356 & \textbf{0.8229} & -2.7656 & 0.7312 & \textbf{0.8508} & -2.0218 & 0.6240 \\

\midrule

\multirow{8}{*}{Gravitational potential}
& MSE &\(\phi\)   & 2  & 21.8936 & 42.7973 & \textbf{12.2100} & 33.1469 & 59.7601 & \textbf{22.7342} & 69.5432 & 79.8920 & \textbf{57.3158} \\
& MSE &\(\phi\)   & 4  & 13.6742 & 42.7643 & \textbf{12.1284} & \textbf{21.8765} & 55.0472 & 22.9810 & 48.1163 & 90.5300 & \textbf{47.6755} \\
& MSE &\(\phi\)   & 8  & 12.4196 & 33.4881 & \textbf{12.0125} & \textbf{22.1266} & 45.9237 & 23.3257 & 53.8123 & 87.1841 & \textbf{49.1723} \\
& MSE &\(\phi\)   & 16 & 12.9412 & 21.7382 & \textbf{12.2687} & \textbf{21.5327} & 31.5565 & 22.0622 & 47.8105 & 61.8841 & \textbf{41.5585} \\
\addlinespace

& \(R^2\) &\(\phi\)   & 2  & 0.9344 & 0.8709 & \textbf{0.9623} & 0.8942 & 0.8162 & \textbf{0.9289} & 0.7815 & 0.7547 & \textbf{0.8198} \\
& \(R^2\) &\(\phi\)   & 4  & 0.9586 & 0.8710 & \textbf{0.9626} & \textbf{0.9320} & 0.8308 & 0.9281 & 0.8515 & 0.7272 & \textbf{0.8534} \\
& \(R^2\) &\(\phi\)  & 8  & 0.9622 & 0.8981 & \textbf{0.9630} & \textbf{0.9278} & 0.8583 & 0.9272 & 0.8312 & 0.7301 & \textbf{0.8471} \\
& \(R^2\) &\(\phi\)   & 16 & 0.9604 & 0.9341 & \textbf{0.9624} & \textbf{0.9323} & 0.9016 & 0.9308 & 0.8515 & 0.8104 & \textbf{0.8744} \\

\hline \hline
\end{tabular}
\end{adjustbox}
\end{table*}

Table~\ref{tab:synthetic_only_equation_recovery} reports equation recovery when SR is trained entirely on generated sequences, without retained real frames. This experiment isolates whether generated sequences alone contain enough physical structure to support downstream SR. For heat conduction, PhysLDM strongly outperforms LDM for all three SR models. The corrected samples reduce \(q\)-MSE by several orders of magnitude, raise \(q\)-\(R^2\) close to one. This confirms that the physics corrector does not only improve field-level fidelity, but also makes the synthetic heat-conduction data usable for recovering Fourier's law. For Navier--Stokes, the synthetic-only results following the same pattern, where PhysLDM improves over LDM for all SR models.

\begin{table*}[!htbp]
\centering
\caption{Synthetic-only equation recovery performance. Results are computed on fully synthetic sequences generated either by the LDM or by the physics-corrected model (PhysLDM). Lower MSE is better, and higher \(R^2\) is better. Bold indicates the best value within each physical dynamics and SR model.}
\label{tab:synthetic_only_equation_recovery}
\small
\begin{adjustbox}{max width=\textwidth}
\begin{tabular}{lllcc}
\hline \hline
\textbf{Physical dynamics} & \textbf{SR model} & \textbf{Variant} & \textbf{MSE} & \(\boldsymbol{R^2}\) \\
\hline \hline

\multirow{6}{*}{Heat conduction}
& DEAP & LDM & 0.1345 & -0.1419\\
& DEAP & PhysLDM & \textbf{0.000456} & \textbf{0.9976} \\
\addlinespace
& GPLearn & LDM & 0.0781 & -0.2197 \\
& GPLearn & PhysLDM & \textbf{0.0000551} & \textbf{0.9938} \\
\addlinespace
& PySR & LDM & 0.1063 & 0.0797 \\
& PySR & PhysLDM & \textbf{0.000461} & \textbf{0.9969} \\

\midrule

\multirow{6}{*}{Navier--Stokes}
& DEAP & LDM & 0.1080 & 0.7901 \\
& DEAP & PhysLDM & \textbf{0.1004} & \textbf{0.8143} \\
\addlinespace
& GPLearn & LDM & 0.1031 & 0.8031 \\
& GPLearn & PhysLDM & \textbf{0.0962} & \textbf{0.8216} \\
\addlinespace
& PySR & LDM & 0.1495 & 0.6945 \\
& PySR & PhysLDM & \textbf{0.1012} & \textbf{0.8116} \\

\midrule

\multirow{6}{*}{Gravitational Potential}
& DEAP & LDM & 72.3799 & 0.7809 \\
& DEAP & PhysLDM & \textbf{22.6559} & \textbf{0.9294} \\
\addlinespace
& GPLearn & LDM & 58.9475 & 0.8226 \\
& GPLearn & PhysLDM & \textbf{12.3912} & \textbf{0.9620} \\
\addlinespace
& PySR & LDM & 110.7011 & 0.6588 \\
& PySR & PhysLDM & \textbf{48.8877} & \textbf{0.8531} \\

\hline \hline
\end{tabular}
\end{adjustbox}
\end{table*}

Table~\ref{tab:fidelity_physical_validity} summarizes the main field-fidelity and physical-validity metrics for the generated data. These metrics are not identical across physical dynamics, because each system has different physically meaningful quantities. Heat conduction is evaluated using flux, reconstructed temperature, divergence, and Fourier-law residual metrics. Navier--Stokes is evaluated using packed-field fidelity, incompressibility, momentum residuals, boundary consistency, and pressure regularity. The gravitational-potential track is evaluated using motion consistency, trajectory reconstruction, spectral structure, and global physical quantities. The fidelity results show that the physics-informed corrector consistently improves physically meaningful quantities. For heat conduction, PhysLDM improves all reported field-level and physics-consistency metrics. The reconstructed-temperature MSE decreases from \(0.00831\) to \(0.000140\), a \(98.31\%\) reduction, and the Fourier residual RMSE decreases from \(2.0056\) to \(0.0738\), a \(96.32\%\) reduction. For Navier--Stokes, PhysLDM improves the packed-field error metrics and gives a clearer advantage for the physical validity metrics. The momentum residual loss decreases from \(3.1864\) to \(0.2826\), a \(91.13\%\) reduction, the divergence loss decreases by \(65.90\%\), and the pressure-in-solid loss is eliminated. For the gravitational-potential track, PhysLDM reduces the constant-velocity residual by \(99.97\%\), the trajectory RMSE by \(98.64\%\), and the RSPS log-MSE by \(99.12\%\). It also improves all reported global physical quantities, including fitted mass, fitted speed, energy, and field statistics.

\begin{table*}[!htbp]
\centering
\caption{Generated-data fidelity and physical validity metrics. GT/reference values are included where applicable. Lower is better for error and residual metrics, while higher is better for \(R^2\) and correlation metrics. Bold indicates the better generated result between LDM and PhysLDM.}
\label{tab:fidelity_physical_validity}
\scriptsize
\begin{adjustbox}{max width=\textwidth}
\begin{tabular}{lllcccc}
\hline \hline
\textbf{Physical dynamics} & \textbf{Metric group} & \textbf{Metric} & \textbf{GT/reference} & \textbf{LDM} & \textbf{PhysLDM} & \textbf{PhysLDM Error Reduction vs LDM} \\
\hline \hline

\multirow{6}{*}{Heat conduction} 
& \multirow{6}{*}{Field fidelity} 
& Face-flux MSE & 0.0000 & 0.00856 & \textbf{0.000508} & 94.06\% \\ 
& & Face-flux \(R^2\) & 1.0000 & 0.8831 & \textbf{0.9925} & 93.58\% \\
& & Center-flux \(q\)-MSE & 0.0000 & 0.00852 & \textbf{0.000499} & 94.14\% \\ 
& & Center-flux \(q\)-\(R^2\) & 1.0000 & 0.8835 & \textbf{0.9927} & 93.69\% \\ 
& & Reconstructed-temperature MSE & 0.0000 & 0.00831 & \textbf{0.000140} & 98.31\% \\ 
& & Reconstructed-temperature \(R^2\) & 1.0000 & 0.8942 & \textbf{0.9981} & 98.16\% \\ 

\midrule
\multirow{3}{*}{Heat conduction}
& \multirow{3}{*}{Physics consistency} 
& \(\nabla \cdot \mathbf{q}\) MSE & 0.0000 & 0.000288 & \textbf{0.0000302} & 89.54\% \\ 
& & Fourier residual \(L^1\) & 0.01470 & 0.5867 & \textbf{0.0431} & 92.66\% \\ 
& & Fourier residual RMSE & 0.03313 & 2.0056 & \textbf{0.0738} & 96.32\% \\ 

\midrule

\multirow{1}{*}{Navier--Stokes}
& Field fidelity & Masked MSE & 0.0000 & 0.00130 & \textbf{0.000992} & 23.48\% \\ 
\midrule
\multirow{6}{*}{Navier--Stokes}
& \multirow{6}{*}{Physical validity} 
& Divergence loss & \(3.51\cdot 10^{-7}\) & 0.0389 & \textbf{0.0133} & 65.90\% \\ 
& & Momentum residual loss & 0.2270 & 3.1864 & \textbf{0.2826} & 91.13\% \\ 
& & Boundary-condition loss & \(1.37\cdot 10^{-16}\) & 0.0509 & \textbf{0.00825} & 83.80\% \\ 
& & Pressure-gauge loss & \(2.58\cdot 10^{-16}\) & \(3.12\cdot 10^{-5}\) & \textbf{\(3.46\cdot 10^{-17}\)} & \(99.99\%\) \\ 
& & Pressure-in-solid loss & 0.0000 & 0.9425 & \textbf{0.0000} & 100.00\% \\ 
& & Mean solid-face velocity & 0.0000 & 0.1524 & \textbf{0.0860} & 43.55\% \\ 

\midrule

\multirow{3}{*}{Gravitational potential}
& \multirow{3}{*}{Motion fidelity}
& Constant-velocity residual & 0.0000 & 105.0885 & \textbf{0.0287} & 99.97\% \\ 
& & Trajectory RMSE & 0.0000 & 4.7450 & \textbf{0.0647} & 98.64\% \\ 
& & Velocity RMSE & 0.0000 & 0.3792 & \textbf{0.0039} & 98.96\% \\ 
\midrule
\multirow{2}{*}{Gravitational potential}
& \multirow{2}{*}{Spectral fidelity} 
& RSPS log-MSE to GT & 0.0000 & 0.4463 & \textbf{0.0039} & 99.12\% \\ 
& & Trajectory-spectrum correlation to GT & 1.0000 & 0.9995 & \textbf{1.0000} & 100\% \\ 
\midrule
\multirow{7}{*}{Gravitational potential}
& \multirow{7}{*}{Physical quantities} 
& Energy error & 0.0000 & 0.4332 & \textbf{0.0125} & 97.11\% \\ 
& & Fitted mass error & 0.0000 & 6.5643 & \textbf{0.1637} & 97.51\% \\ 
& & Fitted speed error & 0.0000 & 0.4264 & \textbf{0.0007} & 99.83\% \\ 
& & Field mean error & 0.0000 & 0.0074 & \textbf{0.0022} & 70.16\% \\ 
& & Field standard-deviation error & -- & 0.2133 & \textbf{0.0101} & 95.28\% \\ 
& & Field minimum error & 0.0000 & 1.8597 & \textbf{0.0888} & 95.23\% \\ 
& & Field maximum error & 0.0000 & 0.0506 & \textbf{0.0233} & 53.99\% \\ 

\hline \hline
\end{tabular}
\end{adjustbox}
\end{table*}

\section{Discussion}
\label{sec:discussion}
In this study, we proposed a physics-informed diffusion-based DE framework for down-the-line SR scientific discovery tasks. We evaluate the proposed framework on several spatiotemporal physical dynamics and out-of-the-box SR models. Across these settings, the results clearly show that synthetic data enrichment is beneficial only when the generated data preserves the physical structure required by the downstream SR task.

Specifically, Table~\ref{tab:sparsity_equation_discovery} shows that physics-corrected data enrichment is most useful when real observations are highly sparse. For heat conduction, Real+PhysLDM consistently gives the best performance across all sparsity levels and SR models, indicating that the corrector restores the flux structure needed to recover Fourier's law. This agrees with previous physics-informed equation-learning studies showing that incorporating physical structure can make model discovery more reliable when observations are scattered, sparse, or noisy \cite{reinbold2021robust,raissi2019physics}. The relatively poor performance of Real+LDM, especially for heat conduction and Navier--Stokes, further supports the view that simply increasing the amount of data is insufficient for equation discovery unless the added data preserve the relevant physical structure. This is consistent with prior PDE-discovery work showing that equation learning is sensitive to data quality, derivative estimation, noise, and incomplete observations \cite{lagergren2020learning,reinbold2020noisy}. For Navier--Stokes, the benefit of Real+PhysLDM is strongest at \(N=2\) and \(N=4\), while Real-only becomes stronger at \(N=8\) and \(N=16\). This agrees with previous findings that derivative-based discovery benefits strongly from accurate observations, since errors in temporal or spatial derivatives can propagate into the recovered equation  \cite{lagergren2020learning,reinbold2020noisy}. For the gravitational-potential task, Real+PhysLDM improves performance for GPLearn and PySR across all sparsity levels, but DEAP remains more competitive with Real-only at higher \(N\). This is in line with symbolic-regression benchmarking studies showing that different SR algorithms can exhibit different sensitivities to data distributions, search spaces, and model-complexity trade-offs \cite{zegklitz2021benchmarking,LaCava2021SRBench}.

Table~\ref{tab:synthetic_only_equation_recovery} further shows that fully synthetic data can support equation recovery, but primarily when the generated sequences are physically corrected. For heat conduction and gravitational potential, PhysLDM strongly outperforms LDM across all three SR models, showing that the corrected generated samples contain enough physically meaningful structure to recover the target equations even without retained real frames. For Navier--Stokes, PhysLDM also improves over LDM for all three SR models, although the gains are more modest than in the heat-conduction and gravitational-potential tracks. In particular, PhysLDM improves both MSE and \(R^2\) for DEAP, GPLearn, and PySR, indicating that physics correction still improves the equation-recovery usefulness of fully synthetic flow sequences. These results agree with recent constrained and physics-guided diffusion studies, which show that adding constraints during sampling or correction can move generated samples toward physically admissible regions rather than only toward statistically plausible samples \cite{christopher2024projected,yuan2022physdiff}. They also agree with physics-constrained symbolic-regression work showing that physical principles can enable robust model discovery from noisy, incomplete, and high-dimensional data \cite{reinbold2021robust}. At the same time, the different improvement magnitudes across physical systems and SR backends suggest that physics correction improves generated data for downstream SR, but does not affect every target equation and regression backend equally. Such SR model dependence agrees with previous SR benchmarking results, where algorithmic choices and search strategies lead to different behavior across problem classes \cite{zegklitz2021benchmarking,LaCava2021SRBench}.

Finally, Table~\ref{tab:fidelity_physical_validity} explains why the downstream SR results differ between uncorrected and physics-corrected generations. Across all three physical dynamics, PhysLDM improves the reported field-fidelity and physical-validity metrics, including Fourier residuals for heat conduction, divergence and momentum residuals for Navier--Stokes, and trajectory, spectral, and fitted-quantity errors for the gravitational-potential system. These results agree with prior work on physics-guided and constrained generative models, where physical projections or constraint-aware sampling improve the physical plausibility of generated outputs \cite{christopher2024projected,yuan2022physdiff}. They also agree with PDE-discovery studies showing that preserving derivative structure and physical consistency is important for reliable recovery of governing equations \cite{long2019pdenet2,reinbold2020noisy}. Taken together, the three result tables support the main conclusion that synthetic data enrichment improves SR only when the generated data are not merely statistically plausible, but also consistent with the physical relations that the downstream SR model is expected to recover.

From an applicative point of view, the proposed framework can be introduced to existing SR-based discovery systems to further improve their performance. In particular, the proposed framework is beneficial for spatiotemporal physical tasks where data is scarce or challenging to produce and obtain as indicated by the obtained results. That said, the advantage of the proposed method is inherently dependent on the quality of the physical information introduced to it, which may result in worse results in downstream SR tasks when performed poorly \cite{wang2022l2loss,kuang2024structural}. 

This study is not without limitations. First, the experiments are based on synthetic data, where the physical parameters and conditioning variables are known. This makes it possible to evaluate equation recovery in a controlled way, but it does not include all sources of uncertainty present in real experimental measurements \cite{alkhalifah2022mlreal,tremblay2018training}. Second, the current SR setup is condition-matched: synthetic completions are generated using the same physical configuration as the corresponding ground-truth sequence. This evaluates sparse-sequence completion, but does not by itself establish performance under extrapolation to unseen physical regimes. Finally, the current top-up protocol does not explicitly enforce temporal consistency across the interfaces between retained real frames and generated frames. Thus, a possible extension of the proposed framework would be to use anchor-conditioned in-between generation, where the model is explicitly conditioned on the retained real frames and trained to generate temporally consistent intermediate states between them. Such a formulation would better match the sparse-frame equation-discovery task than generating a full rollout and then splicing retained ground-truth frames into the sequence. Moreover, future work will benefit from examining whether the same pattern holds under stronger distribution shifts, noisier observations, and more difficult symbolic targets.

Taken jointly, our results indicate that physics-informed generative LDMs can improve downstream SR models' performance in a consistent manner. These findings suggest that synthetic data are most useful for scientific discovery when they are not only plausible samples from the data distribution, but also consistent with the governing relations that downstream inference methods are designed to recover.

\section*{Declarations}
\subsection*{Funding}
None.

\subsection*{Conflicts of interest/Competing interests}
None.

\subsection*{Data and code availability}
All the data and code used for this study is freely available at: \url{https://github.com/simondereuver/PGDM}

\subsection*{AI usage}
The authors used AI-based tools for initial ideations, problem definition, code generation, and the initial version of the manuscript. All materials were manually reviewed and edited. The authors take full responsibility for the content. 

\subsection*{Author Contribution}
Simon De Reuver:  Methodology, Software, Validation, Formal analysis, Investigation, Data Curation, Writing - Original Draft, Writing - Review \& Editing, Visualization\\  
Tamas Kristof Toth:  Methodology, Software, Validation, Formal analysis, Investigation, Data Curation, Writing - Original Draft, Writing - Review \& Editing, Visualization\\  
Teddy Lazebnik: Conceptualization, Writing - Original Draft, Writing - Review \& Editing, Resources, Visualization, Supervision, Project administration\\

\bibliography{biblio}
\bibliographystyle{unsrt}

\end{document}